%% file: main.tex
\documentclass[a4paper]{article}
\usepackage{INTERSPEECH2020}
\usepackage{hyperref}
\usepackage{mathtools}
\usepackage{comment}
\usepackage{amsmath}
\usepackage{CJKutf8}
\usepackage{supertabular}
\usepackage{multirow}
\usepackage{diagbox}
\usepackage{enumerate}

\DeclareMathOperator*{\argmin}{arg\,min}

\title{g2pM: A Neural Grapheme-to-Phoneme Conversion Package for Mandarin Chinese Based on a New Open Benchmark Dataset}

\name{Kyubyong Park\textsuperscript{*}\thanks{\textsuperscript{*}Equal contribution.}, Seanie Lee\textsuperscript{*}}
\address{
  $^1$Author Affiliation\\
  $^2$Co-author Affiliation}
\address{
  $^1$Kakao Brain, South Korea\\
  $^2$KAIST, South Korea}
\email{kyubyong.park@kakaobrain.com, lsnfamily02@kaist.ac.kr}

\begin{document}
\begin{CJK*}{UTF8}{gbsn}

\maketitle

\input{0abstract}


\input{1introduction}
\input{2relatedwork}

\input{3polyphones}
\input{4dataset}
\input{5method}
\input{6experiments}
\input{7g2pM}

\input{8conclusion}

\bibliographystyle{IEEEtran}

\bibliography{main}

\clearpage

\end{CJK*}
\end{document}

%% file: 0abstract.tex
\begin{abstract}

Conversion of Chinese graphemes to phonemes (G2P) is an essential component in Mandarin Chinese Text-To-Speech (TTS) systems. 
One of the biggest challenges in Chinese G2P conversion is how to disambiguate the pronunciation of polyphones---characters having multiple pronunciations. 
Although many academic efforts have been made to address it, there has been no open dataset that can serve as a standard benchmark for a fair comparison to date. 
In addition, most of the reported systems are hard to employ for researchers or practitioners who want to convert Chinese text into pinyin at their convenience.
Motivated by these, in this work, we introduce a new benchmark dataset that consists of 99,000+ sentences for Chinese polyphone disambiguation. 
We train a simple Bi-LSTM model on it and find that it outperforms other pre-existing G2P systems and slightly underperforms pre-trained Chinese BERT. Finally, we package our project and share it on PyPi.
\end{abstract}

%% file: 1introduction.tex
\section{Introduction}
Chinese grapheme to phoneme (G2P) conversion is a task that changes Chinese text into pinyin, an official Romanization system of Chinese. 
It is considered essential in Chinese Text-to-Speech (TTS) systems as unlike English alphabets, Chinese characters represent the meanings, not the sounds. 
A major challenge in Chinese G2P conversion is how to disambiguate the pronunciation of polyphones---characters having more than one pronunciation. In the example below, the first 的\: is pronounced \emph{de}, which means the possessive particle ``of", while the second one is pronounced \emph{dì}, which denotes the ``purpose". 

\begin{itemize}
\item
\emph{Input:}
今天来\textbf{的}目\textbf{的}是什么？ \\
\emph{Translation:}
What is the purpose of coming today? \\
\emph{Output:}
jīn tiān lái \textbf{de} mù \textbf{dì} shì shén me ?
\end{itemize}


There have been many academic efforts to tackle this problem \cite{Shan2016ABL, lz11, lqtzs10, Cai2019PolyphoneDF, hjws, zme02, h08}. However, we find there exist two main problems with them. First, there are no standard benchmark datasets for Chinese polyphone disambiguation. As shown in Table \ref{datasource}, most past works collect copyright data from the Internet, and annotate themselves. Due to the lack of a public benchmark dataset, they report results on different datasets. This makes it hard to compare different models. Second, all of the reports in Table \ref{datasource} do not lead to the release of source code or packages where researchers or practitioners can convert Chinese text into pinyin at their convenience. 

Motivated by these, we construct and release a new Chinese polyphone dataset and a Chinese G2P library using it. 
Our contribution is threefold:
\begin{itemize}
    \item We create a new Chinese polyphonic character dataset, which we call  \textbf{C}hinese \textbf{P}olyphones with \textbf{P}inyin (CPP).
    It is freely available via our GitHub repository\footnote{\url{https://github.com/kakaobrain/g2pM}}.
    \item With the CPP dataset, we train simple neural network models for the Chinese polyphonic character to pinyin task. We find that our best model outperforms other existing G2P systems.
    \item We build a user-friendly Chinese G2P Python library based on one of our models, and share it on PyPi.
    
\end{itemize}

\begin{table}[t]
    \centering
    \small
    \begin{tabular}{p{4.2mm}p{3.8mm}p{29.8mm}p{12.mm}p{5mm}}
    \toprule
	\textbf{Work} & \textbf{Year}      & \textbf{Data Source}     & \textbf{License} & \textbf{Code}          \\
    \midrule
    {\cite{hjws}}  & {2001} & {Ren Ming Daily}  & {copyright} & {N/A}\\
    {\cite{zme02}} & {2002} & {People Daily} & {copyright} & {N/A}\\
    {\cite{h08}} & {2008}& {Sinica and China Times} & {copyright} & {N/A} \\
    {\cite{yjjx09}} & {2009} & {People’s Daily} & {copyright} & {N/A}\\
    {\cite{lqtzs10}} & {2010} & {People’s Daily} & {copyright} & {N/A}\\
    {\cite{lz11}} & {2011} & {People’s Daily} & {copyright} & {N/A}\\
    {\cite{Dong2004GraphemetophonemeCI}} & {2004}  & {People Daily} & {copyright} &{N/A}\\
    {\cite{Shan2016ABL}} & {2016} & {the Internet} & {copyright} & {N/A}\\
    \cite{Cai2019PolyphoneDF} & {2019} & {Data Baker Ltd} & {copyright} & {N/A}\\
    \bottomrule
    \end{tabular}
    \caption{Summary of major past works. Note that most of them source the data from the Internet news articles so it is impossible to access. \cite{Cai2019PolyphoneDF} use a commercial company's internal dataset which is not freely available.}
    \vspace{-0.1in}
	\label{datasource}
\end{table}
\vspace{-0.1in}

%% file: 2relatedwork.tex
\section{Related Work}
\textbf{G2P} There are several works for Chinese polyphone disambiguation. They can be categorized into the traditional rule-based approach \cite{hjws, zme02, h08} and the data driven approach \cite{Shan2016ABL, lz11, lqtzs10,Cai2019PolyphoneDF, mao2007inequality}. The rule-based approach chooses the pronunciation of the polyphonic character based on predefined complex rules along with a dictionary. However, this requires a substantial amount of linguistic knowledge. 
The data driven approach, by contrast, adopts statistical methods such as Decision Tree \cite{lqtzs10} or Maximum Entropy Model \cite{lz11, mao2007inequality}. Recently  \cite{Shan2016ABL, Cai2019PolyphoneDF} use bidirectional Long Short-Term Memory (LSTM) \cite{lstm}  to extract diverse features on the character, word, and sentence level. However, as they depend on external tools such as a word segmenter and a Part-Of-Speech tagger which are not perfect, they are inherently prone to the cascading errors. Recently, some works \cite{mg2p, nmt} consider graphemes to phonemes as sequence transduction and leverage   encoder-decoder architecture to generate multilingual phonemes.

\noindent
\textbf{Benchmark Dataset} To the best of out knowledge, there are no standard benchmark datasets for Chinese polyphone disambiguation. In contrast, there are several public benchmark datasets for English G2P such as CMUDict, Pronlex and NetTalk.

%% file: 3polyphones.tex
\section{Chinese Characters and Polyphones}
We explore what percentage of Chinese characters are polyphones to gauge how important the polyphone disambiguation task is in Chinese.

We  download  the  latest  Chinese  wiki  dump  file\footnote{\url{https://dumps.wikimedia.org/zhwiki/latest/zhwiki-latest-pages-articles.xml.bz2}} and extract plain Chinese text with WikiExtractor \footnote{\url{https://github.com/attardi/wikiextractor}}. All characters including white spaces except Chinese characters are removed. As shown in Table \ref{tab:poly}, the remaining text consists of 17,720 unique characters, or 363M character instances. 
Meanwhile, we collect the list of polyphones from the open-source dictionary, CC-CEDICT\footnote{\url{https://cc-cedict.org/wiki/}}. According to it, 762 out of the 17,720 characters, which account for only 4.30\%, turn out to be polyphones. However, they  occur 67M times in the text, accounting for as much as 18.49\%. This indicates that disambiguating polyphones is a serious problem in Chinese. The most frequent 100 polyphones and their frequencies are provided in Appendix  for reference.

\begin{table}
	\small
	\centering
	\begin{tabular}{llll}
	    \toprule
		{} & {\textbf{Total}} & {\textbf{Monophones}} & {\textbf{Polyphones}}  \\
		\midrule[0.8pt]
		{\# unique char.} & {17,720} & {16,929 (95.60\%)} & {762 (4.30\%)} \\
		{\# characters} & {363M} & {296M (81.51\%)} & {67M (18.49\%)} \\
		
		\bottomrule[0.8pt]
	\end{tabular}
	\caption{Percentage of Chinese polyphones in Wikipedia. A monophone is a character that has a single pronunciation.}
    \vspace{-0.1in}
	\label{tab:poly}
\end{table}

%% file: 4dataset.tex
\begin{figure}[t]
    \centering
    \includegraphics[width=1.0\linewidth]{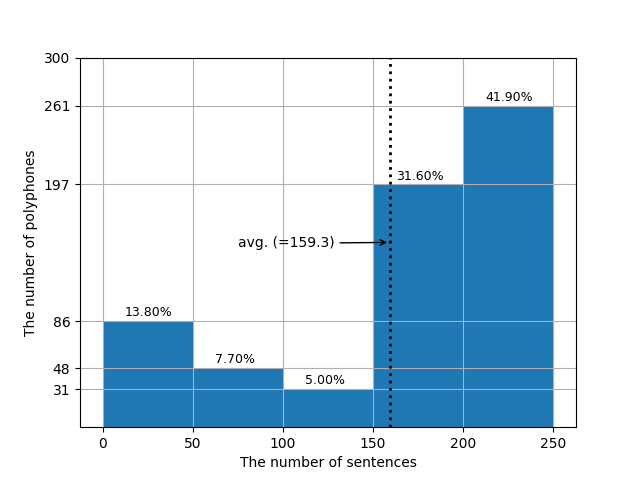}
    \caption{The number of sentences for each polyphonic characters in CPP dataset. On average, a polyphonic character has about 159 sentences.}
    \label{fig:poly-stat}
\end{figure}

\begin{table}[t]
	\small
	\centering
	\begin{tabular}{lllll}
	    \toprule
		{} & {\textbf{Total}} & {\textbf{Train}} & {\textbf{Dev.}} & {\textbf{Test}} \\
		\midrule[0.8pt]
		{\# sentences} & {99,264} & {79,117} & {9,893} & {10,254}\\
		{\# characters per sent} & {31.30} & {31.29} & {31.24} & {31.43}\\
        {\# polyphones}  &  {623} & {623} & {623} & {623} \\
		\bottomrule[0.8pt]
	\end{tabular}
	\caption{Basic statistics of CPP dataset}
    \vspace{-0.1in}
	\label{tab:stat}
\end{table}

\begin{table}[t]
	\small
	\centering
	\begin{tabular}{lll}
	    \toprule
		{\textbf{\# Pronunciations}} & {\textbf{\# Polyphones}} & {\textbf{\# Sentences}} \\
		\midrule[0.8pt]
		{Total} & {623 (100\%)} & {99,264 (100\%)} \\
		{2} & {553 (88.8\%)} & {87,584 (88.2\%)} \\
        {3}  &  {60 (9.6\%)} & {10,162 (10.2\%)} \\
        {4-5}  &  {10 (1.6\%)} & {1,518 (1.6\%)} \\
		\bottomrule[0.8pt]
	\end{tabular}
	\caption{The number of polyphones and sentences in the CPP dataset by the number of possible pronunciations}
    \vspace{-0.1in}
	\label{tab:stat2}
\end{table}

\section{The CPP (\underline{C}hinese \underline{P}olyphones with \underline{P}inyin) Dataset}

In this section, we introduce the CPP dataset---a new Chinese polyphonic character dataset for the polyphone disambiguation task.

\subsection{Data Collection}
We split the aforementioned Chinese text in Wikipedia into sentences. If a sentence contains any traditional Chinese characters, it is filtered out. Also, sentences whose length is more than 50 characters or less than 5 characters are excluded. 
Then, we leave only the sentences having at least one polyphonic character. A special symbol \_ (U+2581) is added to the left and right of a polyphonic character randomly chosen in a sentence to mark the target polyphone. Finally, in order to balance the number of samples across the polyphones, we clip the minimum and maximum number of sentences for any polyphones to 10 and 250, respectively. 
 
\subsection{Human Annotation}
We have two native Chinese speakers annotate the target polyphonic character in each sentence with appropriate pinyin. To make it easier, we provide them with a set of possible pronunciations extracted from CC-CEDICT for the polyphonic character. Next, we ask the annotators to choose the correct one among those candidates. It is worth noting that we do not split the data in half for assignment. Instead, we assign both of the annotators the same entire sentences. Then, we compare each of their annotation results, and discard the sentence if they do not agree.
 
 \subsection{Data Split}
As a result, 99,264 sentences, each of which includes a target polyphone with the correct pinyin annotation, remain. 
Subsequently, we group them by polyphones. For each group, we shuffle and split the sentences into training, development, and test sets at the ratio of 8:1:1. See Table \ref{tab:stat} for details. An example whose target polyphone is 角~ and its correct pinyin is `jiao3' is shown below, where the digit denotes Chinese phonetic tone.

\begin{itemize}
\item
\emph{Sentence:}
即闽粤赣三\_角\_地带。 \\
\emph{Label:}
jiao3 \\
\end{itemize}

 \subsection{Statistics}
Figure \ref{fig:poly-stat} shows how many sentence samples each of the polyphones in the CPP dataset has. 73.5\% of polyphones (458 of 623) have 150-250 samples, while only 13.8\%, i.e., 86 polyphones have less than 50 samples. Obviously, this comes from the differences in the frequency of polyphones.

We also present how many pronunciations the polyphones in the dataset can have in Table \ref{tab:stat2}. Among 623 polyphones in the dataset, 553 (88.8\%) have two possible pronunciations. There are 60 (9.6\%) polyphones in the dataset that can have three pronunciations, and the rest 10 can have up to five pronunciations. All things being equal, we suppose the more pronunciations a polyphone can have, the more challenging it is for a predictor to disambiguate its correct pronunciation. 

\begin{figure}[t]
    \centering
    \includegraphics[width=1.0\linewidth]{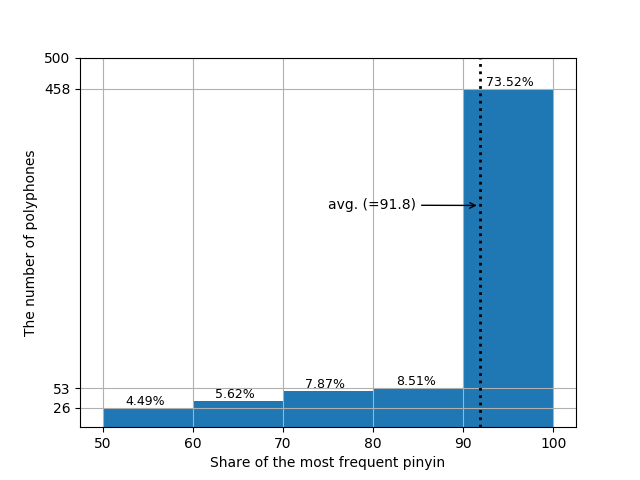}
    \caption{The number of polyphones by the share of the most frequent pinyin for each polyphonic character.}
    \vspace{-0.1in}
    \label{fig:majority-ratio}
\end{figure}

Finally, we explore how dominant the most frequent pronunciation in each polyphone is. As shown prominently in Figure \ref{fig:majority-ratio}, 73.52\% of polyphones are associated with a single prevalent pronunciation that accounts for more than 90\% of all samples. This implies that majority vote---picking up the pronunciation that occurrs most frequently in the training set---would be a strong baseline. However, it is also important to remember there are still many that are less inclined to a dominant pronunciation so majority vote is less effective. 

%% file: 5method.tex
\begin{figure}[t]
	\begin{center}
		\includegraphics[height=5cm]{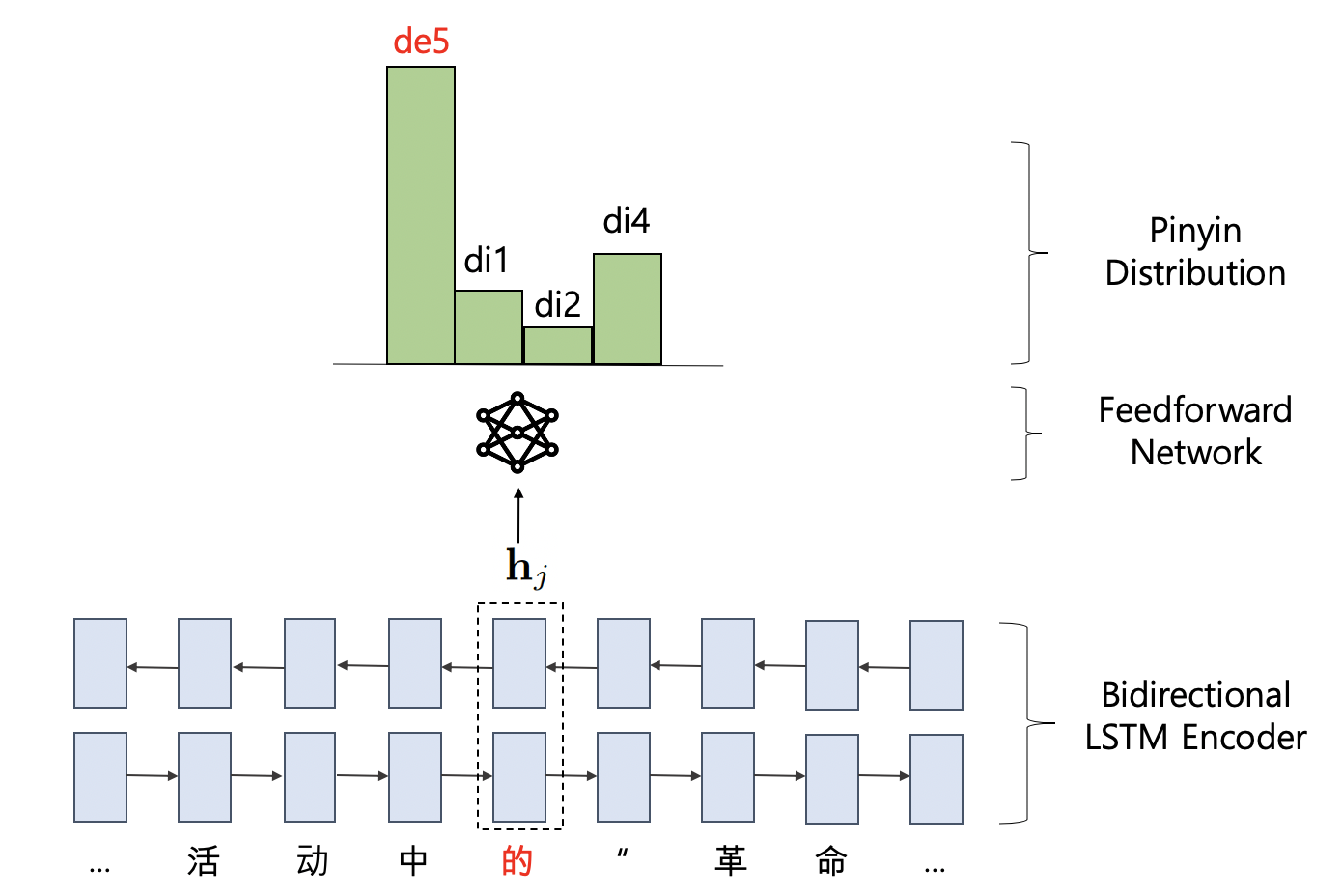}
	\end{center}
	\caption{Conceptual illustration of our models. A sequence of dense character embeddings are encoded with bidirectional LSTMs and the hidden state of the polyphonic character (red-colored) is fed to the feedforward network. It outputs the distribution of the pinyin candidates and finally the most probable one, ``de5" here, is decided as the pronunciation of the character 的.}
	\vspace{-0.1in}
	\label{fig:concept}
\end{figure}

\section{Method}
We consider Chinese polyphone disambiguation as a classification problem and train a function, parameterized by neural networks, which maps a polyphonic character to its pronunciation. 

We do not use any external language processing tools such as word segmenter, entity recognizer, or Part-Of-Speech tagger. Instead, we take as input a sequence of characters and train the network in the end-to-end manner.

\subsection{Embedding}
Let $\mathbf{x} = (x_1, \ldots, x_{T}) \in \mathbb{R}^{T}$ a sequence of characters, which represent a  sentence. We map each character $x_t$ to the dense embedding vector $\mathbf{e}_t \in \mathbb{R}^d$ with a randomly initialized lookup matrix $\mathbf{E} \in \mathbb{R}^{\mathcal{V} \times d}$, where $\mathcal{V}$ is the number of all characters and ${d}$ is the dimension of the embedding vectors. We denote a sequence of character embedding vectors by $\mathbf{e} = (\mathbf{e}_1, \ldots, \mathbf{e}_{T} )$.

\subsection{Bidirectional LSTM Encoder}
The bidirectional Long Short-Term Memory (Bi-LSTM) \cite{lstm} network is used to encode the contextual information of the polyphonic character. 
At any time step $t$, the representation $\mathbf{h}_t$ is the concatenation of the forward hidden state 

($\overrightarrow{\mathbf{h}}_t$) and the backward hidden state ($\overleftarrow{\mathbf{h}}_t$).
\begin{align*}
\begin{split}
    \overrightarrow{\mathbf{h}}_t &= \overrightarrow{\text{LSTM}}(\mathbf{e_t}, \overrightarrow{\mathbf{h}}_{t-1}) \\
    \overleftarrow{\mathbf{h}}_t &= \overleftarrow{\text{LSTM}}(\mathbf{e}_t, \overleftarrow{\mathbf{h}}_{t-1}) \\
    \mathbf{h}_t &= concat(\overrightarrow{\mathbf{h}}_t,\overleftarrow{\mathbf{h}}_t) 
\end{split}
\end{align*}

\subsection{Fully Connected Layers}
We use two fully connected layers to transform the encoded information into the classification label. Let $j$ the position index of the polyphonic character in the sentence. The concatenated hidden state $\mathbf{h}_j$ (dotted line in Figure \ref{fig:concept}) is fed into the two-layered feedforward network followed by the softmax function, yielding the pinyin probability distribution $\mathbf{\hat{y}}$ over all possible pinyin classes as follows:
\begin{align}
    \mathbf{\hat{y}} = (\hat{y}_1, \ldots, \hat{y}_c) = \text{softmax}(g_2(\varphi(g_1(\mathbf{h}_j))))
\end{align}
where $g_1$ and $g_2$ are fully connected layers, and $\varphi$ is a non-linear activation function such as ReLU \cite{relu}, and $c$ is the number of possible pinyin classes.

\subsection{Loss Function}
Let $\mathbf{y} = (y_1,\ldots, y_c) \in \mathbb{R}^c$ be a one-hot vector of a true label. We use cross-entropy as a loss function for training. In other words, we minimize the negative log-likelihood to find the optimal parameters $\theta$, which we denote as $ \hat{\theta}$.
\begin{align}
    \mathcal{L}(\theta)& = -\sum\limits_{j=1}^c y_j \log (\hat{y_j}) \\
    \hat{\theta} &= \argmin_{\theta} \mathcal{L}(\theta)
\end{align}

%% file: 6experiments.tex
\section{Experiments}

\subsection{Training}
We randomly initialize the character embedding matrix and fix its dimension to 64. 
To find the optimal hyperparameter values, we vary the hidden size\footnote{The hidden size in this context refers to the size \emph{after} the concatenation of the forward and backward hidden states.} in (16, 32, 64) and the number of layers in the Bi-LSTM encoder in (1, 2, 3). The dimension of the last two fully connected layers is set to 64, and ReLU \cite{relu} is used as the activation function.We train all the models with Adam optimizer \cite{adam} and batch size 32 for 20 epochs. All the experiments are run five times with different random seeds.

\begin{table}[t]
	\centering
	\small
	\begin{tabular}{cccc}
	    \toprule
		\diagbox[width=1.0cm]{\textbf{H}}{\textbf{L}}  & {1} & {2} & {3} \\
		\midrule[0.8pt]
		{16} & {$94.34 \pm 0.17$} & {$92.99 \pm 0.11$} & {$85.30 \pm 4.50$}\\
		{32} & {$96.64 \pm 0.04$} & {$96.01 \pm 0.14 $} & {$95.75 \pm 0.14$} \\
		{64} & {$\bm{97.15 \pm 0.09}$} & {$97.09 \pm 0.05$} & {$96.58 \pm 0.07$}\\
		\bottomrule[0.8pt]
	\end{tabular}
	\caption{Development set accuracy of varying models by the hidden size (denoted as \textbf{H}) and the number of LSTM layers (denoted as \textbf{L}). Note that the model in bold face is the best one.}
	\vspace{-0.1in}
	\label{tab:exp2}
\end{table}

\begin{table}[t]
	\small
	\centering
	\begin{tabular}{lc}
	    \toprule
		{\textbf{System}} & {\textbf{Test Accuracy}} \\
		\midrule[0.8pt]
		{Majority vote} & {92.08}\\
		{xpinyin (0.5.6)} & {78.56} \\
		{pypinyin (0.36.0)} & {86.13} \\
		{g2pC (0.9.9.3)} & {84.45} \\
		{Chinese BERT} & \textbf{97.85} \\
		{Ours} & {97.31} \\
		\bottomrule[0.8pt]
	\end{tabular}
	\caption{Test set accuracy of Chinese g2p systems}
	\vspace{-0.1in}
	\label{tab:exp}
\end{table}
\subsection{Evaluation}

\noindent
\textbf{Hyperparameter Search} Table \ref{tab:exp2} summarizes the development set accuracy of various models according to the hidden size and the number of layers in the Bi-LSTM encoder. We observe that the bigger the hidden size is, the higher the accuracy is, as expected. However, we get the better result when we use the fewer number of layers. The model of a single layer with 64 hidden units shows the best performance.

\noindent
\textbf{Baseline \& other systems} As we mentioned earlier, we take so-called ``majority vote'' as a baseline. 
It decides the pronunciation of a polyphonic character by simply choosing the most frequent one in the training set.
For example, 咯\: can be pronounced \emph{luò}, \emph{gē}, and \emph{lo},  and their frequencies in the CPP training set are 63, 51, and 2, respectively. 
At test time, the majority vote system always picks up \emph{luò} for 咯, irrespective of the context.

We also compare our model with three open-source Chinese G2P libraries: xpyinin\footnote{\url{https://github.com/lxneng/xpinyin}}, pyinin\footnote{\url{https://github.com/mozillazg/python-pinyin}}, and g2pC\footnote{\url{https://github.com/Kyubyong/g2pC}}. 
xpinyin and pypinyin are based on rules, while g2pC uses Conditional Random Fields (CRFs)\cite{crf} for polyphone disambiguation. All of them are easily accessible through PyPi.

Finally, we test the pretrained Chinese BERT model~\cite{cbert}. We take a finetuning approach; we attach a fully connected layer to the BERT network and feed the hidden state of the polyphonic character to it. We do not freeze any weights.

\noindent
\textbf{Results} Our model slightly underperforms Chinese BERT and outperforms all the other systems by large margin.
As shown in Table \ref{tab:exp}, ours reaches 97.31\% accuracy on the test set, which is 4.33\% point higher than the majority vote and 0.54 lower than Chinese BERT.
That our simple neural model shows comparable performance to the heavy BERT model, which has more than 102M parameters, tells us two things. One is that our model is simple but powerful enough. Another is that it is not too simple for the naïve majority vote to beat.

%% file: 7g2pM.tex
\section{g2pM: a Grapheme-to-Phoneme Conversion Library for Mandarin Chinese}
We develop a simple Chinese G2P library in Python, dubbed \emph{g2pM}, using our best Bi-LSTM model. The package provides an easy-to-use interface in which users can convert any Chinese sentence into a list of the corresponding pinyin. We share it on PyPi at \url{https://pypi.org/project/g2pM/}.

\subsection{Packaging}
We implement g2pM purely in Python. In order to minimize the number of external libraries that must be pre-installed, we first re-write our Pytorch inference code in NumPy \cite{scipy}.
Our best model is 1.7MB in size, and the package size is a little bigger, 2.1MB, as it includes some contents of CC-CEDICT. Details are shown in Table \ref{tab:setup}. g2pM works like the following. Given a Chinese text, g2pM checks every character if it is a polyphonic character. If so, the neural network model returns its predicted pronunciation. Otherwise, the pronunciation of the (monophonic) character is retrieved from the dictionary contained in the package.

\begin{table}[t]
	\small
	\centering
	\begin{tabular}{lc}
	    \toprule
		{\textbf{Layer}} & {\textbf{Size}}  \\
		\midrule[0.8pt]
		{Embedding} & {64}  \\
		{LSTM $\times$1} & {64} \\
		{Fully Connected $\times$2} & {64} \\
		\midrule[0.8pt]
		{Total \# parameters} & {477,228} \\
		{Model size} & {1.7MB} \\
		{Package size} & {2.1MB} \\
		\bottomrule[0.8pt]
	\end{tabular}
	\caption{Breakdown of g2pM. $\times$ denotes the number of layers.}
	\vspace{-0.1in}
	\label{tab:setup}
\end{table}

\begin{figure}[h]
    \centering
    \includegraphics[width=1.0\linewidth]{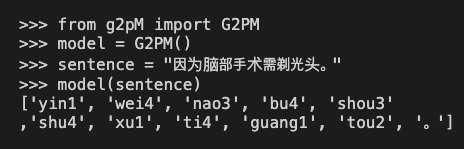}
    \caption{Usage example of g2pM.}
    \vspace{-0.1in}
    \label{fig:code}
\end{figure}

\subsection{Usage}
g2pM provides simple APIs for operation.
With a few lines of code, users can convert any Chinese text into a sequence of pinyin. An example is available in Figure \ref{fig:code}. More details are on the Github repository.

%% file: 8conclusion.tex
\section{Conclusion}
We proposed a new benchmark dataset for Chinese polyphone disambiguation, which is freely and publicly available. We trained simple deep learning models, and created a Python package with one of them. We hope our dataset and library will be helpful for researchers and practitioners.